\documentclass[11pt,a4paper]{article}
\usepackage[hyperref]{acl2021}
\usepackage{times}
\usepackage{latexsym}

\usepackage{microtype}

\usepackage{CJKutf8}
\usepackage{multirow}
\usepackage{graphicx}
\usepackage{colortbl}
\usepackage{arydshln}
\usepackage{makecell}
\usepackage{booktabs}
\usepackage{amsmath}
\usepackage{amssymb}
\usepackage{diagbox}

\newcommand{\CalD}{\mathcal{D}} 
\newcommand{\CalS}{\mathcal{S}}
\newcommand{\CalL}{\mathcal{L}}

\newcommand{\Vh}{\boldsymbol{\mathit{h}}}

\newcommand{\SetD}{\mathbb{D}}
\newcommand{\SetS}{\mathbb{S}}
\newcommand{\SetP}{\mathbb{P}}

\newcommand{\Mh}{\boldsymbol{\mathit{H}}}
\newcommand{\Mx}{\boldsymbol{\mathit{X}}} 
\newcommand{\My}{\boldsymbol{\mathit{Y}}} 

\usepackage{xspace}
\newcommand{\ke}{{\textsc{Ke}}\xspace}
\newcommand{\ts}{{\textsc{Ts}}\xspace}
\newcommand{\rd}{{\textsc{Rd}}\xspace}
\newcommand{\sys}{{\textsc{All}}\xspace}
\newcommand{\dialogpt}{DialoGPT\xspace}

\aclfinalcopy 
\def\aclpaperid{2188} 

\title{Language Model as an Annotator: Exploring DialoGPT \\ for Dialogue Summarization}

\author{
Xiachong Feng$^{1}$, 
Xiaocheng Feng$^{1,2}$\Thanks{~Corresponding author.},
Libo Qin$^{1}$, 
Bing Qin$^{1,2}$, 
Ting Liu$^{1,2}$ \\
$^{1}$Harbin Institute of Technology, China\\
$^{2}$Peng Cheng Laboratory, China\\
\texttt{\{xiachongfeng,xcfeng,lbqin,bqin,tliu\}@ir.hit.edu.cn}
}

\date{}

\begin{document}
\maketitle
\begin{abstract}
Current dialogue summarization systems usually encode the text with a number of general semantic features (e.g., keywords and topics) to gain more powerful dialogue modeling capabilities.
However, these features are obtained via open-domain toolkits that are dialog-agnostic or heavily relied on human annotations.
In this paper, we show how DialoGPT \cite{dialogpt}, a pre-trained model for conversational response generation, can be developed as an unsupervised dialogue annotator, which takes advantage of dialogue background knowledge encoded in DialoGPT.
We apply DialoGPT to label three types of features on two dialogue summarization datasets, SAMSum and AMI, and employ pre-trained and non pre-trained models as our summarizers.
Experimental results show that our proposed method can obtain remarkable improvements on both datasets and achieves new state-of-the-art performance on the SAMSum dataset\footnote{Our codes are available at: \url{https://github.com/xcfcode/PLM_annotator}}.
\end{abstract}

\section{Introduction}

Dialogue summarization aims to generate a succinct summary while retaining essential information of the dialogue \cite{gurevych-strube-2004-semantic,chen2020multiviewsm}.
Theoretically, \newcite{peyrard-2019-simple} point out that a good summary is intuitively related to three aspects, including {\em Informativeness}, {\em Redundancy} and {\em Relevance}.

To this end, previous works have taken the above three aspects into account by incorporating auxiliary annotations into the dialogue.
To improve informativeness, some works annotated linguistically specific words (e.g., nouns and verbs), domain terminologies and topic words in the dialogue \cite{Riedhammer2008AKB,Koay2020HowDT,Zhao2020ImprovingAD}.
To reduce redundancy, some works used sentence similarity-based methods to annotate redundant utterances.
\cite{Zechner2002AutomaticSO,Murray2005ExtractiveSO}. 
To improve relevance, some works annotated topics for the dialogue \cite{Li2019KeepMS,Liu2019TopicAwarePN,chen2020multiviewsm}.
However, these annotations are usually obtained via open-domain toolkits, which are not suitable for dialogues, or require manual annotations, which are labor-consuming.

To alleviate the above problem, we explore the pre-trained language model as an unsupervised annotator to automatically provide annotations for the dialogue.
Recently, some works have investigated the use of pre-trained language models in an unsupervised manner. 
For example, \newcite{Sainz2021Ask2TransformersZD} exploited pre-trained models for assigning domain labels to WordNet synsets.
The successful recipe is that a model is obtained extensive knowledge via pre-training on a huge volume of data.
When it comes to the dialogue domain, \dialogpt \cite{dialogpt} is a SOTA conversational response generation model, which is pre-trained on the massive dialogue data. 
Therefore, we draw support from \dialogpt and present our {\it \dialogpt annotator}, which can perform three dialogue annotation tasks, including keywords extraction, redundancy detection and topic segmentation, to measure informativeness, redundancy and relevance of the input dialogue, respectively.

\begin{figure*}[t]
	\centering
	\includegraphics[scale=0.41]{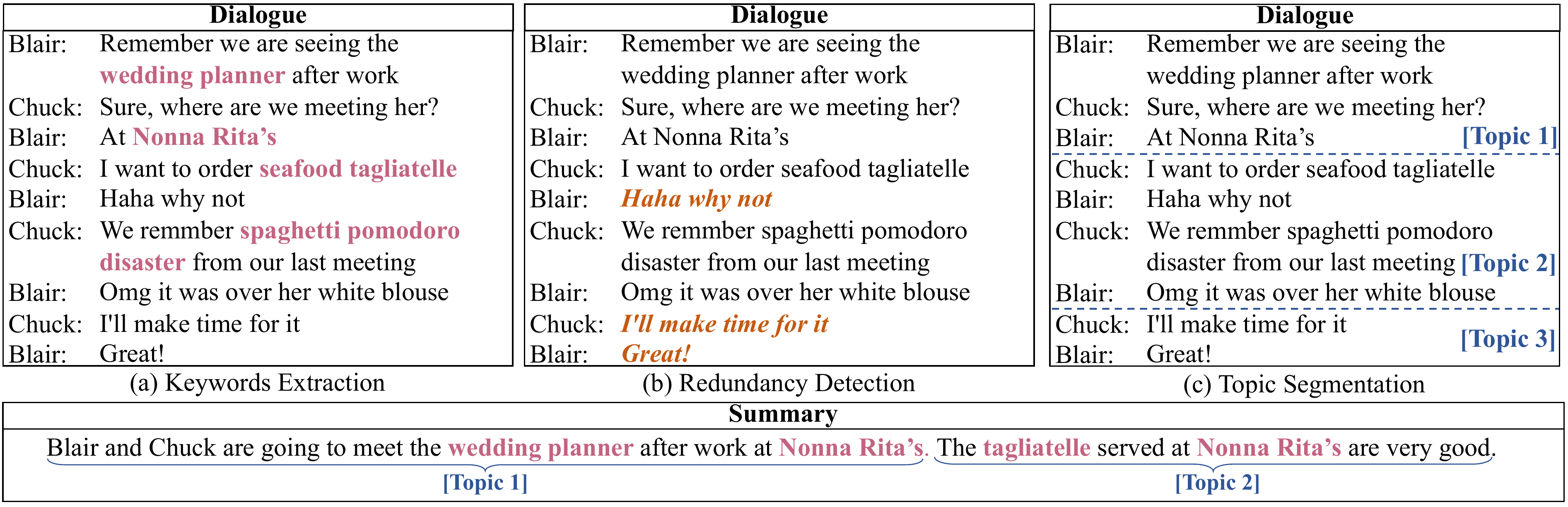}
	\caption{Example dialogue from SAMSum \cite{samsum} with the human annotated summary. (a) Keywords extraction aims to extract words that are most important to the dialogue. (b) Redundancy detection aims to detect nonsignificant utterances in the dialogue. (c) Topic segmentation aims to divide the whole dialogue into several fine-grained topics. All three auxiliary information can do good to final summary generation.}
	\label{fig:intro}
\end{figure*}

\textbf{Keywords Extraction} aims to automatically identify important words in the dialogue (shown in Figure \ref{fig:intro}(a)).
Our {\it \dialogpt annotator} extracts unpredictable words as keywords.
We assume that keywords contain high information, which are difficult to be predicted considering both background knowledge encoded in the \dialogpt and contextual information of dialogue context.
\textbf{Redundancy Detection} aims to detect redundant utterances that have no core contribution to the overall meaning of the dialogue (shown in Figure \ref{fig:intro}(b)).
Our {\it \dialogpt annotator} detects utterances that are useless for dialogue context representation as redundant.
We assume that if adding a new utterance does not change the dialogue context representation, then this utterance has no effect on predicting the response, so it is redundant.
\textbf{Topic Segmentation} aims to divide a dialogue into topically coherent segments (shown in Figure \ref{fig:intro}(c)).
Our {\it \dialogpt annotator} inserts a topic segmentation point before one utterance if it is unpredictable.
We assume that if an utterance is difficult to be inferred from the dialogue context based on \dialogpt, this utterance may belong to a new topic.

We use our {\it \dialogpt annotator} to annotate the SAMSum \cite{samsum} and AMI \cite{ami} datasets.
Each annotation is converted into a specific identifier and we  insert them into the dialogue text.
Then, we employ pre-traind BART \cite{bart} and non pre-trained PGN \cite{pgn} as our summarizers. 
Extensive experimental results show that our method can obtain consistent and remarkable improvements over strong baselines on both datasets and achieves new state-of-the-art performance on the SAMSum dataset.

\section{Preliminaries}
In this section, we will describe the task definition as well as the background of \dialogpt.

\subsection{Task Definition}
Given an input dialogue ${\CalD}$, a dialogue summarizer aims to produce a condensed summary ${\CalS}$, where ${\CalD}$ consists of $|{\CalD}|$ utterances $[u_1,u_2,...u_{|{\CalD}|}]$ and ${\CalS}$ consists of $|{\CalS}|$ words $[s_1,s_2,...s_{|{\CalS}|}]$. Each utterance $u_i$ is compose of a sequence of words $[u_{i,1},u_{i,2},...u_{i,|u_i|},\texttt{EOS}_i]$, where $i \in [1:|{\CalD}|]$ and $\texttt{EOS}_i$ indicates the end of the utterance. Besides, each utterance $u_i$ associates with a speaker $p_i$. Thus, this task can be formalized as producing the summary ${\CalS}$ given the dialogue sequence: $\CalD = [p_1, u_{1,1},...,\texttt{EOS}_1 ,...,p_{|\CalD|},u_{|\CalD|,1},...,\texttt{EOS}_{|\CalD|}]$

\begin{figure*}[t]
	\centering
	\includegraphics[scale=0.47]{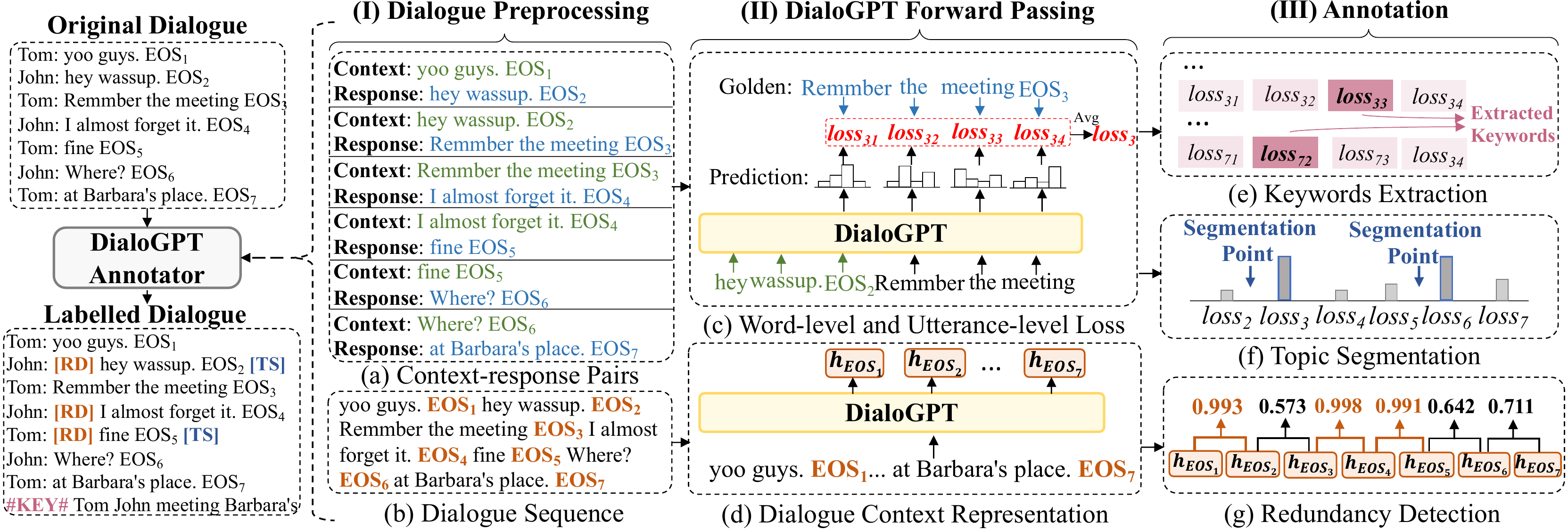}
	\caption{Illustration of our \dialogpt annotator. (\uppercase\expandafter{\romannumeral1}) Given one dialogue, we preprocess it into two formats: context-response pairs and the dialogue sequence. (\uppercase\expandafter{\romannumeral2}) We input them into the DialoGPT, after the forward pass, we can get the word-level and utterance-level predicted losses and representations for dialogue context. (\uppercase\expandafter{\romannumeral3}) We perform three annotation tasks: keywords extraction, redundancy detection and topic segmentation. Finally, we can get a labelled dialogue. {\texttt{\#KEY\#}}, {\texttt{[RD]}} and {\texttt{[TS]}} are specific tags, which are inserted into the dialogue.}
	\label{fig:framework}
\end{figure*}

\subsection{\dialogpt}
\dialogpt \cite{dialogpt} is a neural conversational response generation model, which inherits from GPT-2 \cite{gpt2} and is trained on 147M conversation-like exchanges extracted from Reddit comment chains. There are 3 different sizes of the model with total parameters of 117M, 345M and 762M respectively. It achieves state-of-the-art results over various dialogue generation benchmarks. Given the dialogue context $u_{i-1}=[u_{i-1,1},...,u_{i-1,|u_{i-1}|},\texttt{EOS}_{i-1}]$, \dialogpt aims to produce the response $u_i=[u_{i,1},...,u_{i,|u_{i}|},\texttt{EOS}_{i}]$, which can be formalized as the conditional probability of $P(u_{i}|u_{i-1})$. It first takes the context word sequence of no more than 1024 tokens and outputs the representation of the sequence $h_i=({\Vh}_{i-1,1},...,{\Vh}_{i-1,|u_{i-1}|},{\Vh}_{i-1,\texttt{EOS}_{i-1}})$, where ${\Vh}_{i-1,\texttt{EOS}_{i-1}}$ can be viewed as the representation of dialogue context $u_{i-1}$.  Then, \dialogpt starts decoding the response by attending to the context token representations and partially decoded response tokens until reaching \texttt{EOS}. The loss function is the negative log-likelihood of the response word sequence ${\CalL}_{\dialogpt}=-\sum_{t=1}^{|u_i|} \log p\left(u_{i,t} | u_{i,1} \ldots u_{i,t-1}, u_{i-1}\right)$. It's worth noting that \dialogpt tokenizes texts with the same byte-pair encoding as GPT-2, thus either context or response tokens are tokenized into subwords.

\section{Method}

In this section, we will first introduce our \dialogpt annotator. The workflow consists of three steps (1) dialogue preprocessing; (2) \dialogpt forward passing; (3) annotation. The overall framework is shown in Figure \ref{fig:framework}. Then, we will describe our dialogue summarizer, including BART and PGN.

\subsection{Dialogue Preprocessing}
Dialogue preprocessing aims to transform the original dialogue  $\CalD = [p_1, u_{1,1},...,\texttt{EOS}_1,...,p_{|\CalD|},u_{|\CalD|,1},...,\texttt{EOS}_{|\CalD|}]$  into the format that \dialogpt can process. 

Specifically, we transform it into two formats.
The first one is \textbf{context-response pairs} (shown in Figure \ref{fig:framework}(a)).
Given a dialogue $\CalD$, two adjacent utterances  $(u_{i-1},u_i)$ are combined into a context-response pair, where $i \in [2:|\CalD|]$ .
The second one is \textbf{dialogue sequence} (shown in Figure \ref{fig:framework}(b)).
All the utterances in the dialogue $\CalD$ are serialized into a sequence $[u_{1,1},...,\texttt{EOS}_1,...,u_{|\CalD|,1},...,\texttt{EOS}_{|\CalD|}]$, with \texttt{EOS} separates each utterance.

Note that either for context-response pairs or the dialogue sequence, we do not take speaker information $p$ into consideration. 
The reason is that \dialogpt is trained on a huge volume of  conversational data without speaker information.
Even so, \newcite{dialogpt} proved that \dialogpt can simulate real-world dialogues in various scenes and has already learned diverse response generation patterns between the same speakers or different speakers according to the given context.

\subsection{\dialogpt Forward Passing}
\dialogpt forward passing has two purposes.
(1) For each context-response pair, we aim to get the word-level and utterance-level predicted losses for the response (shown in Figure \ref{fig:framework}(c)).
(2) For the dialogue sequence, we aim to get the representations for each \texttt{EOS} (shown in Figure \ref{fig:framework}(d)).

For the first purpose, given one context-response pair $(u_{i-1},u_i)$, we input the context words $u_{i-1}=[u_{i-1,1},u_{i-1,2},...,u_{i-1,|u_{i-1}|}, \texttt{EOS}_{i-1}]$ into the \dialogpt and start to decode the response. At each decode step $t$, we calculate the negative log-likelihood between the predicted distribution and the golden target from the given response. 
\begin{equation}
\setlength{\abovedisplayskip}{3pt}
\setlength{\belowdisplayskip}{3pt}
\begin{split}
    & loss_{i,t}=-\log p\left(u_{i,t} | {u_{i,<t}}, u_{i-1}\right) \\
    & loss_{i}=\frac{1}{|u_i|+1} \sum_{t=1}^{|u_i|+1}loss_{i,t}
\end{split}
\end{equation}
where $loss_{i,t}$ and $loss_{i}$ are the predicted losses for each word and each utterance respectively\footnote{Note that \dialogpt uses BPE to tokenize texts, thus, losses are calculated at the sub-word level. We recover the word-level predicted loss by averaging the losses of multiple sub-words. Besides, since the first utterance $u_1$ can only be served as the context, so we do not compute loss for $u_1$.}.

For the second purpose, after the single forward pass of \dialogpt over the dialogue sequence, we can get representations $\Mh$ for each token on the top of the \dialogpt. Afterward, we extract all representations for each \texttt{EOS}.
\begin{equation}
\setlength{\abovedisplayskip}{3pt}
\setlength{\belowdisplayskip}{3pt}
\begin{split}
    &\Vh_{\texttt{EOS}_1},\Vh_{\texttt{EOS}_2},...,\Vh_{\texttt{EOS}_{|\CalD|}} = \Mh(\texttt{EOS})
\end{split}
\end{equation}
where each $\Vh_{\texttt{EOS}_i}$ can be viewed as the representation for the dialogue context $[u_1,...,u_i]$.

\subsection{Annotation}

\begin{figure}[t]
	\centering
	\includegraphics[scale=0.19]{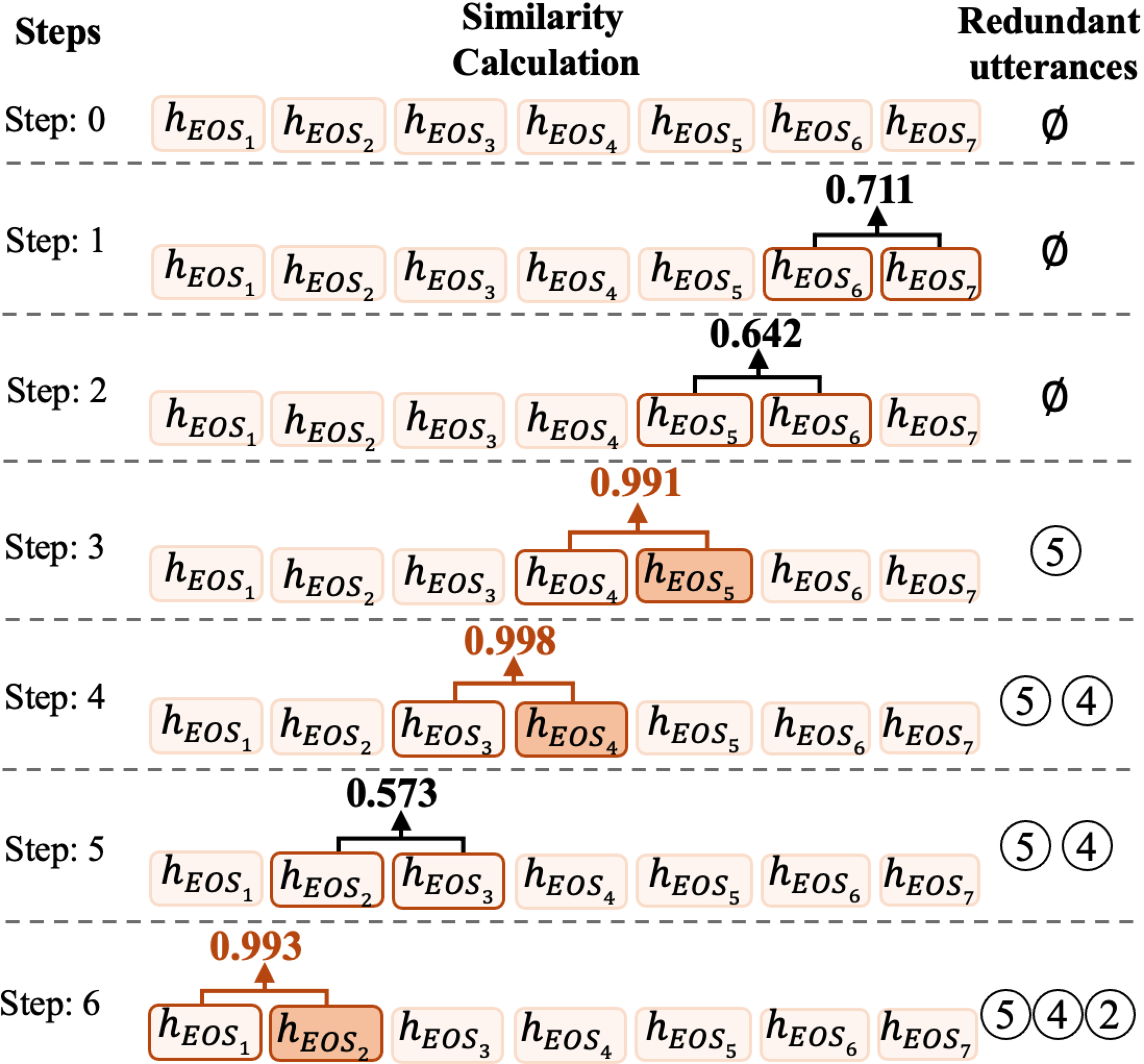}
	\caption{Illustration of redundancy detection process. The initial redundant utterances set is $\emptyset$. $\Vh_{\texttt{EOS}_i}$ is the representation for dialogue context covering the first $i$ utterances. We detect redundant utterances based on the cosine similarity between representations of dialogue context. For example, the similarity score between $\Vh_{\texttt{EOS}_4}$ and $\Vh_{\texttt{EOS}_5}$ exceeds the pre-defined threshold ($t_{\rd}$ is 0.99), which means adding utterance $u_5$ into the dialogue context brings little information, thus the utterance $u_5$ is detected as redundant.
 }
	\label{fig:rd}
\end{figure}

\subsubsection{Keywords Extraction: $\text{\dialogpt}_{\ke}$}\label{sec:ke}
\noindent \textbf{Motivation}
Considering both background knowledge encoded in the \dialogpt and contextual information of the dialogue context, if one word in the golden response is difficult to be inferred from \dialogpt, we assume that it contains high information and can be viewed as a keyword.

Given a dialogue $\CalD$, we have loss $loss_{i,j}$ for each word $u_{i,j}$, where $i \in [2:|\CalD|]$. We extract $r_{\ke}$ percent of words with the highest loss as keywords, where $r_{\ke}$ is a hyper-parameter\footnote{We use a heuristic rule to predetermine the possible value of $r_{\ke}$ by calculating the average of length of summaries (remove stopwords) divided by the length of dialogues in the train set. We search the best $r_{\ke}$ based on the calculated score.}. Moreover, the names of all speakers $\SetP$ mentioned in the dialogue are also added into the keywords set. Finally, we append a specific tag \texttt{\#KEY\#} and the keywords to the end of the original dialogue $\CalD$. The new dialogue with keywords annotation is $\CalD_{\ke}=[\underbrace{p_1, u_{1,1},...,}_{\CalD} \underbrace{\texttt{\#KEY\#},\SetP,\texttt{Key}_1,\texttt{Key}_2,...}_{keywords}]$.\footnote{In experiments, we find that the predicted loss for the first word of each utterance is extremely high, probably due to the first word in the response is the most uncertain and hard to be predicted. Thus, we ignore the first word of each utterance.}

\subsubsection{Redundancy Detection: $\text{\dialogpt}_{\rd}$}\label{sec:rd}
\noindent \textbf{Motivation} \dialogpt inherits a decoder architecture, where one token attends to all previous tokens to aggregate information. Thus, given the representation $\Vh_{\texttt{EOS}_i}$ for each $\texttt{EOS}_i$, it can be viewed as the representation for the dialogue context $[u_1,u_2,...,u_i]$. Adding a new utterance $u_{i+1}$, if the new context representation $\Vh_{\texttt{EOS}_{i+1}}$ is similar to the previous $\Vh_{\texttt{EOS}_i}$, we assume that the new utterance $u_{i+1}$ brings little information and has small effects on predicting the response, thus $u_{i+1}$ becomes a redundant utterance.

We start with the last two dialogue context representations $\Vh_{\texttt{EOS}_{|\CalD|-1}}$ and $\Vh_{\texttt{EOS}_{|\CalD|}}$, and calculate the cosine similarity between them. If the similarity score exceeds the threshold $t_{\rd}$, the utterance $u_{|\CalD|}$ is detected as redundant. $t_{\rd}$ is a hyper-parameter. 
If the similarity score doesn't exceed the threshold $t_{\rd}$, we move forward one step to calculate the similarity between $\Vh_{\texttt{EOS}_{|\CalD|-2}}$ and $\Vh_{\texttt{EOS}_{|\CalD|-1}}$, and repeat the process until reaching $\Vh_{\texttt{EOS}_{1}}$. An example is shown in Figure \ref{fig:rd}. 

We insert a specific tag \texttt{[RD]} before each redundant utterance. For example, if utterance $u_1$ is redundant, the new dialogue with redundant utterances annotation is $\CalD_{\rd}=[p_1,\texttt{[RD]},u_{1,1},...,\texttt{EOS}_1,...,p_{|\CalD|},...,\texttt{EOS}_{|\CalD|}]$.

\subsubsection{Topic Segmentation: $\text{\dialogpt}_{\ts}$}\label{sec:ts}
\noindent \textbf{Motivation} 
\dialogpt is skilled in generating the context-consistent response. Therefore, if the response is difficult to be predicted given the context based on DialoGPT, we assume the response may belong to another topic and there is a topic segmentation between the context and response.

Given a dialogue $\CalD$, we have loss $loss_{i}$ for each utterance $u_i$, where $i \in [2:|\CalD|]$. 
We select $r_{\ts}$ percent of utterances with the highest loss as topic segmentation points.
$r_{\ts}$ is a hyper-parameter\footnote{We use a heuristic rule to predetermine the possible value of $r_{\ts}$ by calculating the average of the number of summary sentences divided by the number of dialogue utterances in the train set. This is based on the observation that each sentence in golden summary tends to correspond to one topic of the dialogue. We search the best $r_{\ts}$ based on the calculated score.}.
Before each selected utterance, we insert a specific tag \texttt{[TS]}.
For example, if there is a segmentation point between utterance $u_1$ and utterance $u_2$, the new dialogue with topic annotation is $\CalD_{\ts}=[p_1,u_{1,1},...,\texttt{EOS}_1,\texttt{[TS]},p_2,u_{2,1},...,\texttt{EOS}_2,...]$.

\subsection{Summarizer}
We employ two kinds of summarizer, one is \textbf{BART} \cite{bart}, which is a Transformer-based model and pre-trained on a huge volume of data. The other one is \textbf{PGN} \cite{pgn}, which is a LSTM-based model. Both models inherit a typical sequence-to-sequence framework, which first encodes the source dialogue $\CalD$ to distributed representations and then generates the target summary $\CalS$ with the decoder. 

\paragraph{\textbf{BART}} BART adopts the Transformer \cite{transformer} as the backbone architecture. It first map the source dialogue into distributed representations, based on which a decoder generates the target sequence:
\begin{equation}
\begin{split}
 \Mx^N &= \textsc{\textbf{Encoder}}(\Mx^0)\stackrel{N}{\underset{n=1}{:=}} \textsc{Ffn}\left(\textsc{Att}(\Mx^{n-1})\right)  \\
  \My^M &= \textsc{\textbf{Decoder}}(\My^0,\Mx^N)\\
  &\stackrel{M}{\underset{m=1}{:=}} \textsc{Ffn}\left(\textsc{Att}\left(\textsc{Att}(\My^{m-1}), \Mx^N\right)\right)
\end{split}
\label{eq:transformer}
\end{equation}
where $\stackrel{N}{\underset{n=1}{:=}}$ denotes $N$ identical encoding layers, $\stackrel{M}{\underset{m=1}{:=}}$ denotes $M$ identical decoding layers, $\Mx^0$ denotes the sum of the word embeddings $\Mx_{\mathrm{emb}}$ and position embeddings $\Mx_{\mathrm{pos}}$ of $\CalD$, $\My^0$ denotes that of the shifted right $\CalS$, $\textsc{Ffn}(\cdot)$ denotes a position-wise feed-forward network, and $\textsc{Att}(\cdot)$ denotes a multi-head attention.
Residual connection \citep{He:2016ib} and layer normalization \citep{Ba:2016vc} are used in each sub-layer, which are suppressed in Equation~\ref{eq:transformer} for clarity. Finally, the output representation $\My^M$ of the decoder is projected into the vocabulary space and the decoder outputs the highest probability token.

\paragraph{\textbf{PGN}} PGN is a hybrid model of the typical Seq2Seq Attention model \cite{nallapati2016abstractive} and Pointer-Network \cite{pointer}. The input dialogue is fed into the LSTM encoder token by token, producing the encoder hidden states. The decoder receives word embedding of the previous word and generates a distribution to decide the target token, retaining decoder hidden states. PGN not only allows to generate from the fixed vocabulary, but also allows to copy from the input tokens. 

\paragraph{\textbf{Training Objective}}
Model parameters $\theta$ are trained to maximize the conditional likelihood of the outputs in a parallel training corpus $({\SetD},{\SetS})$:
\begin{equation}
    \arg\max_{\theta} \sum_{(\CalD, \CalS) \in (\SetD,\SetS)} \log p (\CalS \,|\, \CalD; \theta).
\end{equation}

\section{Experiments}

\begin{table}[t]
\small
\centering
        \begin{tabular}{c|l|ccc}
            \hline 
            & & \textbf{Train} & \textbf{Valid} & \textbf{Test}  \\
            \hline
            \hline
            \multirow{4}{*}{\rotatebox{90}{{SAMSum}}} & \# & 14732 & 818 & 819  \\
            & Avg.Turns  & 11.13  & 10.72 & 11.24 \\
            & Avg.Tokens & 120.26  & 117.46 & 122.71\\
            & Avg.Sum  & 22.81 & 22.80 & 22.47 \\
            \hline
            \hline
            \multirow{4}{*}{\rotatebox{90}{{AMI}}} & \#  &97 &20 &20  \\
            & Avg.Turns  &310.23 &345.70 &324.40 \\
            & Avg.Tokens  &4859.52 &5056.25 &5257.80 \\
            & Avg.Sum  &323.74 &321.25 &328.20 \\
            \hline
        \end{tabular}
\caption{Statistics for SAMSum and AMI datasets. ``\#" means the number of dialogue-summary pairs, ``Avg.Turns", ``Avg.Tokens" and ``Avg.Sum" mean the average number of turns of dialogues, tokens of dialogues and tokens of summaries respectively.}
\label{tab:datasets}
\end{table}

\subsection{Datasets}
We experiment on 2 datasets (statistics in Table \ref{tab:datasets}):

\noindent \textbf{SAMSum} \cite{samsum} is a human-generated dialogue summary dataset, which contains dialogues in various scenes of the real-life.

\noindent  \textbf{AMI} \cite{ami} is a meeting summary dataset.
Each meeting contains four participants and is about a remote control design project.

\subsection{Implementation Details} 
\noindent \textbf{DialoGPT} We initialize DialoGPT with {\em DialoGPT-large}\footnote{https://huggingface.co/transformers}.
For SAMSum, we set keywords extraction ratio $r_{\ke}$ to 15, similarity threshold $t_{\rd}$ to 0.99 and topic segmentation ratio $r_{\ts}$ to 15.
For AMI, $r_{\ke}$ is 4, $t_{\rd}$ is 0.95 and $r_{\ts}$ is 5 \footnote{We show more hyper-parameter search results for SAMSum and AMI datasets in the supplementary file.}.

\noindent \textbf{BART}
We initialize BART with {\em bart.large}\footnote{https://github.com/pytorch/fairseq} .
For fine-tuning on SAMSum, the learning rate is set to 3e-05, the dropout rate is 0.1, the warmup is set to 400. At the test process, beam size is 5, minimum decoded length is 5 and maximum length is 100.

\noindent \textbf{PGN} The word embedding size is set to 300 and initialized with the pre-trained GloVe vector. The dimension of encoder and pointer decoder is set to 200. The dropout is set to 0.5. The learning rate is 0.001. At the test process, beam size is 10, minimum decoded length is 280 and maximum length is 450\footnote{https://github.com/OpenNMT/OpenNMT-py}.

\begin{table}[t]
\centering
        \begin{tabular}{l|lll}
            \hline
            \textbf{Model} & \textbf{R-1} & \textbf{R-2} & \textbf{R-L}   \\
            \hline
            \hline
            \multicolumn{4}{c}{\it Extractive} \\
            \hline
            LONGEST-3  &32.46 &10.27 &29.92 \\
            TextRank  &29.27 &8.02 &28.78 \\
            \hline
            \hline
            \multicolumn{4}{c}{\it Abstractive} \\
            \hline
            Transformer &36.62 &11.18 &33.06  \\
            D-HGN   &42.03 &18.07 &39.56  \\
            TGDGA   &43.11 &19.15 &40.49  \\
            DialoGPT   &39.77 &16.58 &38.42  \\
            MV-BART   &53.42 &27.98 &$\textbf{49.97}^{\dagger\dagger}$  \\
            \hline
            \hline
            \multicolumn{4}{c}{\it Ours} \\
            \hline
            BART   &52.98 &27.67 &49.06  \\
            \hdashline[1pt/3pt]
            BART($\CalD_{\ke}$) &$\textbf{53.43}^{\dagger\dagger}$ &$\textbf{28.03}^{\dagger\dagger}$ &49.93 \\
            BART($\CalD_{\rd}$)  &53.39 &28.01 &49.49 \\
            BART($\CalD_{\ts}$)  &53.34 &27.85 &49.64 \\
            \hdashline[1pt/3pt]
            BART($\CalD_{\sys}$) &$\textbf{53.70}^{\dagger}$ &$\textbf{28.79}^{\dagger}$ &$\textbf{50.81}^{\dagger}$ \\
            \hline 
        \end{tabular}
\caption{Test set results on the SAMSum dataset, where ``R'' is short for ``ROUGE''. BART means fine-tuning BART on the original SAMSum. BART($\CalD_{\ke}$), BART($\CalD_{\rd}$) and BART($\CalD_{\ts}$) represent fine-tuning BART on the SAMSum with keywords, redundancy and topic annotation respectively. $\CalD_{\sys}$ means the SAMSum with all three annotations. $\dagger$ and $\dagger\dagger$ indicate the first-ranked and second-ranked results respectively.} \label{tab:main_results_samsum}
\end{table}

\begin{table}[t]
\centering
        \begin{tabular}{l|lll}
            \hline
            \textbf{Model} & \textbf{R-1} & \textbf{R-2} & \textbf{R-L}   \\
            \hline
            \hline
            \multicolumn{4}{c}{\it Extractive} \\
            \hline
            TextRank &35.19 &6.13 &15.70 \\
            SummaRunner & 30.98 &5.54 &13.91 \\
            \hline
            \hline
            \multicolumn{4}{c}{\it Abstractive} \\
            \hline
            UNS &37.86 &7.84 &13.72 \\
            TopicSeg &$\textbf{51.53}^{\dagger\dagger}$ &12.23 &$\textbf{25.47}^{\dagger}$ \\
            HMNet &$\textbf{52.36}^{\dagger}$ &$\textbf{18.63}^{\dagger}$ &24.00 \\
            \hline
            \hline
            \multicolumn{4}{c}{\it Ours} \\
            \hline
            PGN  &48.34 &16.02 &23.49 \\
            \hdashline[1pt/3pt]
            PGN($\CalD_{\ke}$)  &50.22 &17.74 &24.11 \\
            PGN($\CalD_{\rd}$) &50.62 &16.86 &24.27 \\
            PGN($\CalD_{\ts}$) &48.59 &16.07 &24.05 \\
            \hdashline[1pt/3pt]
            PGN($\CalD_{\sys}$) &50.91 &$\textbf{17.75}^{\dagger\dagger}$  &$\textbf{24.59}^{\dagger\dagger}$ \\
            \hline
        \end{tabular}
\caption{Test set results on the AMI dataset. PGN($\CalD_{\ke}$), PGN($\CalD_{\rd}$) and PGN($\CalD_{\ts}$) represent training PGN on the AMI with keywords, redundancy and topic annotation respectively.} \label{tab:main_results_ami}
\end{table}

\begin{table}[t]
\centering
        \begin{tabular}{l|c||l|c}
            \hline
            \multicolumn{2}{c||}{\bf SAMSum} & \multicolumn{2}{c}{\bf AMI}\\
            \hline
            \textbf{Model} & \textbf{BS} & \textbf{Model} & \textbf{BS} \\
            \hline
            BART   &86.91   &PGN &80.51    \\
            MV-BART  &88.46 &HMNet  & 82.24 \\
            BART($\CalD_{\sys}$)  &\textbf{90.04}  &PGN($\CalD_{\sys}$)  &\textbf{82.76}  \\
            \hline 
        \end{tabular}
\caption{Test set results on the SAMSum and AMI. ``BS" is short for BERTScore.} \label{tab:new_metric}
\end{table}

\subsection{Baselines and Metrics}
For SAMSum, 
\textbf{LONGEST-3} views the first three utterances as the summary.
\textbf{TextRank} \cite{textrank} is a traditional graph-based method.
\textbf{Transformer} \cite{transformer} is a seq2seq method based on full self-attention operations.
\textbf{D-HGN} \cite{Feng2020IncorporatingCK} incorporates commonsense knowledge to help understand dialogues.
\textbf{TGDGA} \cite{Zhao2020ImprovingAD} uses topic words and models graph structures for dialogues.
\textbf{DialoGPT} \cite{dialogpt} means that fine-tuning DialoGPT on the SAMSum.
\textbf{MV-BART} \cite{chen2020multiviewsm} is a BART-based method that incorporates topic and stage information.

For AMI, 
\textbf{SummaRunner} \cite{nallapati2017summarunner} is an extractive method based on hierarchical RNN network.
\textbf{UNS} \cite{shang-etal-2018-unsupervised} is a fully unsupervised and graph-based method.
\textbf{TopicSeg} \cite{Li2019KeepMS} incorporates topics to model the meeting.
\textbf{HMNet} \cite{zhu2020ahn} is a transformer-based method that incorporates POS and entity information and is pre-trained on news summarization dataset.

We adopt ROUGE \cite{rouge} and BERTScore \cite{bert-score} for evaluating our models.

\subsection{Automatic Evaluation}
The results on SAMSum and AMI are shown in Table \ref{tab:main_results_samsum} and \ref{tab:main_results_ami} respectively. 
We can see that using our annotated datasets $\CalD_{\ke}$, $\CalD_{\rd}$ and $\CalD_{\ts}$, both BART and PGN can obtain improvements. Furthermore, our BART($\CalD_{\sys}$) achieves SOTA performance.

For SAMSum, it's worth noting that BART($\CalD_{\ke}$) performs better compared with BART($\CalD_{\rd}$) and BART($\CalD_{\ts}$). 
We attribute this to the fact that keywords can retain essential information for shorter dialogues.
For AMI, PGN($\CalD_{\rd}$) contributes the most, which shows the importance of detecting redundancy in verbose meeting transcripts.
Although HMNet and TopicSeg achieve better scores, HMNet needs  news summarization dataset to pre-train the model and TopicSeg designs complex attention mechanism to incorporate topic information.

In terms of new embedding-based metric BERTScore (shown in Table \ref{tab:new_metric}), our method BART($\CalD_{\sys}$) and PGN($\CalD_{\sys}$) can consistently outperform the baseline models\footnote{Evaluation details are shown in the supplementary file.}.

\subsection{Human Evaluation}
\begin{table}[t]
\centering
        \begin{tabular}{c|llll}
            \hline
            & \textbf{Model} & \textbf{Info.} & \textbf{Conc.} & \textbf{Cov.} \\
            \hline
            \multirow{7}{*}{\rotatebox{90}{{SAMSum}}} & Golden   &4.37   &4.26   &4.27  \\
            \cdashline{2-5} 
            & BART                  & 3.66   & 3.65    & 3.66 \\
            & MV-BART               & 3.85   & 3.76    & 3.88 \\
            & BART($\CalD_{\ke}$)   & 3.88   & 3.77    & 3.79 \\
            & BART($\CalD_{\rd}$)   & 3.74   & $\textbf{3.98 }^{\dagger}$   & 3.89 \\
            & BART($\CalD_{\ts}$)   & $\textbf{3.95}^{\dagger\dagger}$   & 3.76    & $\textbf{4.01}^{\dagger\dagger}$ \\
            & BART($\CalD_{\sys}$)  & $\textbf{4.05}^{\dagger}$   & $\textbf{3.78}^{\dagger\dagger}$    &  $\textbf{4.08}^{\dagger}$\\
            \hline
            \hline
            \multirow{7}{*}{\rotatebox{90}{{AMI}}} & Golden  &4.70   &3.85   &4.35  \\
            \cdashline{2-5} 
            & PGN  &2.92 &3.08   &2.70  \\
            & HMNet &$\textbf{3.52}^{\dagger}$ &2.40   & $\textbf{3.40}^{\dagger}$ \\
            & PGN($\CalD_{\ke}$)  &3.20 &3.08   &3.00  \\
            & PGN($\CalD_{\rd}$)  &3.15 &$\textbf{3.25}^{\dagger}$   &3.00 \\
            & PGN($\CalD_{\ts}$)  &3.05 &$\textbf{3.10}^{\dagger\dagger}$   &$\textbf{3.17}^{\dagger\dagger}$   \\
            & PGN($\CalD_{\sys}$)  &$\textbf{3.33}^{\dagger\dagger}$ &$\textbf{3.25}^{\dagger}$   &3.10 \\
            \hline 
        \end{tabular}
\caption{Human evaluation results. ``Info." is short for informativeness, ``Conc." for conciseness, ``Cov." for coverage. For SAMSum, the inter-annotator agreement (Fleiss' kappa) scores for each metric are 0.46, 0.37 and 0.43 respectively. For AMI, Fleiss' kappa scores are 0.48, 0.40 and 0.41 respectively.} \label{tab:human}
\end{table}

We conduct a human evaluation of the dialogue summary to assess its informativeness, conciseness and coverage. 
Informativeness measures how well the summary includes key information.
Conciseness measures how well the summary discards the redundant information.
Coverage measures how well the summary covers each part of the dialogue.

We randomly sample 100 dialogues (SAMSum) and 10 meetings (AMI)  with corresponding generated summaries to conduct the evaluation. In order to reduce variance caused by humans, we have 4 human evaluators and they were asked to rate each summary on a scale of 1 to 5 (higher is better) for each metric.
The results are shown in Table \ref{tab:human}.

We can see that our method can achieve higher scores in all three metrics.
Especially, combined with $\CalD_{\rd}$, our model can get the best score in conciseness.
Besides, combined with $\CalD_{\ts}$, our model can perform better in coverage. 
However, HMNet gets the best score in informativeness and coverage. 
We argue this is because HMNet forces a minimum summary length of 400. Due to this, it scores the worst in conciseness.
For the AMI, we also find there is still a gap between the scores of generated summaries and the scores of golden summaries, indicating that the AMI is more difficult.

\subsection{Analysis}

\paragraph{Effect of $\text{\dialogpt}_{\ke}$.}

\begin{table}[t]
\centering
        \begin{tabular}{l|ccc}
            \hline
            \textbf{Method} & \textbf{R-1} & \textbf{R-2} & \textbf{R-L}  \\
            \hline
            \hline
            \multicolumn{4}{c}{\textit{Rule-Based Methods}} \\
            \hline
            Entities  &53.36 &27.71 &49.69   \\
            Nouns and Verbs &52.75 &27.48 &48.82  \\
            \hline
            \hline
            \multicolumn{4}{c}{\textit{Traditional Methods}} \\
            \hline
            TextRank &53.29 &27.66 &49.33  \\
            Topic words &53.28 &27.76 &49.59 \\
            \hline
            \hline
            \multicolumn{4}{c}{\textit{Pre-trained Language Model-Based Methods}} \\
            \hline
            KeyBERT & & &   \\
            \ \ w/ BERT emb &52.39 &27.14 &48.52   \\
            \ \ w/ DialoGPT emb &53.14 &27.25 &49.42   \\
            \hline
            \hline
            \multicolumn{4}{c}{\textit{Ours}} \\
            \hline
            $\text{\dialogpt}_{\ke}$ &\textbf{53.43} &\textbf{28.03} &\textbf{49.93}  \\
            \hline 
        \end{tabular}
\caption{Test set results of fine-tuning BART on the SAMSum that is annotated with keywords using various methods. Entities, nouns and verbs are obtained by \newcite{qi2020stanza}. Topic words are obtained by a pre-trained LDA model \cite{xsum-emnlp}. KeyBERT \cite{grootendorst2020keybert} leverages pre-trained language model embeddings to create keywords.} \label{tab:key}
\end{table}

\begin{table}[t]
    \centering
    \begin{tabular}{c|ccc}
        \hline
        \textbf{Method} & \textbf{Precision} & \textbf{Recall} & $\mathbf{F_1}$  \\
         \hline
         TextRank &47.74\% &17.44\% &23.22\% \\
         Entities & \textbf{60.42\%} &17.80\% &25.38\% \\
         $\text{\dialogpt}_{\ke}$ & 33.20\% &\textbf{29.49\%} &\textbf{30.31\% }\\
         \hline
    \end{tabular}
    \caption{Quantitative evaluation for keywords on SAMSum test set by viewing reference summary words as golden keywords.}
    \label{tab:key_f}
\end{table}

To verify the effectiveness of our $\text{\dialogpt}_{\ke}$ method, we fine-tune BART on SAMSum, which is annotated by various keywords extraction methods.
The results are shown in Table \ref{tab:key}. 
We can see that our method achieves higher scores.
The results also show that entities play an important role in the summary generation. 
Besides, combined with \dialogpt embeddings, KeyBERT can get better results.

To give a quantitative evaluation, we view reference summary words as golden keywords and calculate the precision, recall and $F_1$ scores for extracted keywords.
The results are shown in Table \ref{tab:key_f}.
Directly using entities as keywords can get the best precision score. 
However, both \textit{TextRank} and \textit{Entities} perform poorly in recall.
Our method gets the best score in terms of $F_1$ and its advantage is mainly reflected in recall score, which shows our method can extract more diverse keywords.

\paragraph{Effect of $\text{\dialogpt}_{\rd}$.}

\begin{table}[t]
\centering
        \begin{tabular}{l|ccc}
            \hline
            \textbf{Model} & \textbf{R-1} & \textbf{R-2} & \textbf{R-L}   \\
            \hline
            \hline
            \multicolumn{4}{c}{\textbf{SAMSum}} \\
            \hline 
            Rule-based  &53.00 &27.71 &\textbf{49.68} \\
            \hline 
            $\text{\dialogpt}_{\rd}$ &\textbf{53.39} &\textbf{28.01} &49.49 \\
            \hline 
            \hline
            \multicolumn{4}{c}{\textbf{AMI}} \\
            \hline 
            Rule-based &50.19 &16.45 &23.95\\
            \hline 
            $\text{\dialogpt}_{\rd}$ &\textbf{50.62} &\textbf{16.86} &\textbf{24.27}  \\
            \hline 
        \end{tabular}
\caption{Test set results on the SAMSum and AMI datasets that are annotated with redundant utterances. ``Rule-based" indicates annotating utterances that contain no noun, verb and adjective as redundant.} \label{tab:redun}
\end{table}

To verify the effectiveness of our $\text{\dialogpt}_{\rd}$ method, we compare it with a \textit{Rule-based} method \cite{dinarelli2009annotating}, which annotates utterances without noun, verb and adjective as redundant. 
The results are shown in Table \ref{tab:redun}.
We can see that our method performs better.
Especially, our method shows more advantages for long and verbose meeting transcripts in the AMI.

\paragraph{Effect of $\text{\dialogpt}_{\ts}$.}
To verify the effectiveness of our $\text{\dialogpt}_{\ts}$ method, we compare it with the C99 algorithm \cite{choi-2000-c99}, which is a sentence similarity-based segmentation method. \newcite{chen2020multiviewsm} enhance it with BERT \cite{devlin2019bertpo} embeddings.
We further combine the algorithm with DialoGPT embeddings.
The results are shown in Table \ref{tab:topic}.
We can see that our method can get comparable results with the strong baseline C99(w/ DialoGPT emb). 
For AMI, combined with golden topic annotation, PGN can achieve the best result, which shows modeling topics is an essential task for dialogue summarization.

\begin{table}[t]
\centering
        \begin{tabular}{l|ccc}
            \hline
            \textbf{Model} & \textbf{R-1} & \textbf{R-2} & \textbf{R-L}   \\
            \hline
            \hline
            \multicolumn{4}{c}{\textbf{SAMSum}} \\
            \hline 
            C99 & &  &\\
            \ \ w/ BERT emb  &52.80 &27.78 &49.50 \\
            \ \ w/ DialoGPT emb  &53.33 &\textbf{28.04} &49.39 \\
            $\text{\dialogpt}_{\ts}$ &\textbf{53.34} &27.85 &\textbf{49.64} \\
            \hline 
            \hline
            \multicolumn{4}{c}{\textbf{AMI}} \\
            \hline 
            Golden &50.28 &19.73 &24.45 \\
            \hdashline[1pt/3pt]
            C99 & &  &\\
            \ \ w/ BERT emb  &48.53 &15.84 &23.63 \\
            \ \ w/ DialoGPT emb  &\textbf{49.22} &\textbf{16.79} &23.88 \\
            $\text{\dialogpt}_{\ts}$  &48.59 &16.07 &\textbf{24.05} \\
            \hline 
        \end{tabular}
\caption{Test set results on SAMSum and AMI that are annotated with topic segmentation in various methods. C99 \cite{choi-2000-c99} segments dialogues based on inter-sentence similarities. Beside, the AMI has golden topic segmentation annotations.} \label{tab:topic}
\end{table}

\subsection{Case Study}
\begin{figure*}[!htb]
	\centering
	\includegraphics[scale=0.41]{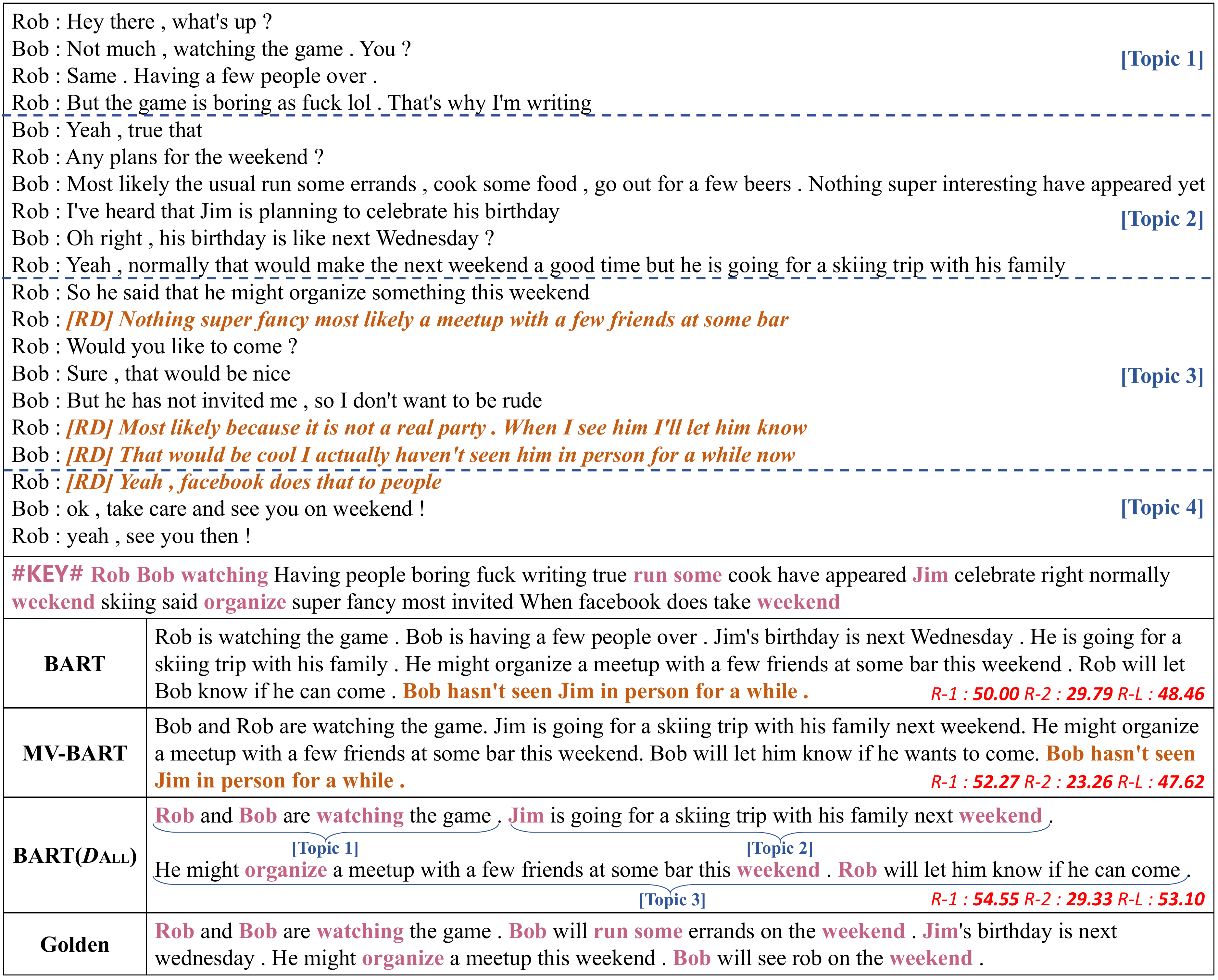}
	\caption{Example dialogue in the SAMSum dataset and summaries generated by different models. Keyowrds, redundant utterances and topics are annotated by our {\it DialoGPT Annotator}. ``R" is short for ROUGE. Our model BART($\CalD_{\sys}$) can get higher ROUGE scores while generating the better summary.}
	\label{fig:case}
\end{figure*}

Figure \ref{fig:case} shows summaries generated by different models for an example dialogue in the SAMSum dataset.
We can see that BART \cite{bart} tends to generate long and redundant summaries.
By incorporating topic and stage information, MV-BART \cite{chen2020multiviewsm} can generate summaries that cover main topics of the dialogue.
However, it still suffers from redundancy problem.
Our BART($\CalD_{\sys}$) can get higher ROUGE scores while generating better summaries.
The generated summary can include extracted keywords and correspond to each topic of the dialogue.
We also find that even some redundant utterances have already been detected, our model still generate the summary contains some redundant information. We attribute this to the fact that the small dataset leads to insufficient training of the model.

\section{Related Work}
\noindent \textbf{Dialogue Summarization}
Current works mainly incorporate auxiliary information to help better modeling dialogues. 
Some works used various types of {\em keywords} to identify the core part of the dialogue, including entities \cite{zhu2020ahn}, domain terminologies \cite{Koay2020HowDT} and topic words \cite{Zhao2020ImprovingAD}.
Some works aimed to reduce {\em redundancy}, \newcite{Zechner2002AutomaticSO,Murray2005ExtractiveSO} used 
sentence-level similarity-based methods.
Some works incorporate {\em topics} as a coarse-grained dialogue structure \cite{Li2019KeepMS,Liu2019TopicAwarePN,chen2020multiviewsm}. 
Other works also explored {\em dialogue act} \cite{Goo2018AbstractiveDS}, {\em dialogue discourse} \cite{Feng2020DialogueDG} and {\em commonsense knowledge} \cite{Feng2020IncorporatingCK}.
In this paper, we combine three types of auxiliary information to help better modeling dialogues, including keywords, redundant utterances and topics.

\noindent \textbf{Pre-trained Language Models}
Pre-trained models such as BERT \cite{devlin2019bertpo} and GPT-3 \cite{Brown2020LanguageMA} have advanced various NLP tasks.
On one hand, some works utilized the knowledge contained in pre-trained models by fine-tuning on supervised data of downstream tasks \cite{Qin2019ASF,Liu2019TextSW,Qin2020CoGATAC}. 
On the other hand, some works examined the knowledge in an unsupervised manner \cite{Jiang2019HowCW,emnlp20-modeling,lin-etal-2020-birds}. \newcite{Kumar2020DataAU} explored pre-trained models for conditional data augmentation. 
\newcite{Wang2020LanguageMA} used the knowledge in pre-trained models to construct knowledge graphs.
In this paper, we belong to the second paradigm and propose our \dialogpt annotator that can perform three annotation tasks in an unsupervised manner.

\section{Conclusion}
We investigate to use \dialogpt as unsupervised annotators for dialogue summarization, including keywords extraction, redundancy detection and topic segmentation.
We conduct our \dialogpt annotator on two datasets, SAMSum and AMI.
Experimental results show that our method consistently obtains improvements upon pre-traind summarizer (BART) and non pre-trained summarizer (PGN) on both datasets.
Besides, combining all three annotations, our summarizer can achieve new state-of-the-art performance on the SAMSum dataset.

\section*{Acknowledgments}
This work is supported by the National Key R\&D Program of China via grant 2018YFB1005103 and National Natural Science Foundation of China (NSFC) via grant 61906053 and 61976073. We thank all the anonymous reviewers for their insightful comments. We also thank Lifu Huang and Xinwei Geng for helpful discussion.

\bibliography{acl2021}
\bibliographystyle{acl_natbib}

\appendix

\section{Evaluation Details}
For ROUGE \cite{rouge}, we employ Py-rouge\footnote{https://pypi.org/project/py-rouge/} package to evaluate our models following \newcite{samsum}. For BERTScore \cite{bert-score}, we use the official implementation\footnote{https://github.com/Tiiiger/bert\_score} to evaluate our models. The detailed command line for BERTScore is \texttt{bert-score -r golden.txt -c gen.txt --lang en}.

\section{Ablation Studies for Annotations}
To further verify the effectiveness of our method, we conduct ablation studies for each annotation. The results are shown in Table \ref{tab:ab_samsum} and Table \ref{tab:ab_ami}.
We can find that: (1) For both datasets, training summarizers based on datasets with two of three annotations can obtain improvements. (2) For both datasets, training summarizers based on datasets with two of three annotations can surpass corresponding summarizers that are trained based on datasets with one type of annotation (e.g.,  BART($\CalD_{\ke+\rd}$) is better than  BART($\CalD_{\ke}$) and  BART($\CalD_{\rd}$)). (3) Compared with summarizers that are trained on $\CalD_{\rd+\ts}$ and $\CalD_{\ke+\rd}$, summarizers that are trained on $\CalD_{\ke+\ts}$ get relatively small improvements on both datasets. Nevertheless, it indicates that $\text{\dialogpt}_{\ke}$ and $\text{\dialogpt}_{\ts}$ still have non-overlapping parts. (4) Combining all three annotations, both summarizers can achieve the best results in all ROUGE scores.

\begin{table}[t]
\centering
        \begin{tabular}{l|lll}
            \hline
            \textbf{Model} & \textbf{R-1} & \textbf{R-2} & \textbf{R-L}   \\
            \hline
            \hline
            \multicolumn{4}{c}{\it Ours} \\
            \hline
            BART   &52.98 &27.67 &49.06  \\
            \hdashline[1pt/3pt]
            BART($\CalD_{\ke}$) &53.43 &28.03 &49.93 \\
            BART($\CalD_{\rd}$)  &53.39 &28.01 &49.49 \\
            BART($\CalD_{\ts}$)  &53.34 &27.85 &49.64 \\
            \hdashline[1pt/3pt]
            BART($\CalD_{\ke+\rd}$) &53.56 &28.65 &50.55 \\
            BART($\CalD_{\ke+\ts}$)  &53.51 &28.13 &50.00 \\
            BART($\CalD_{\rd+\ts}$)  &53.64 &28.33 &50.13 \\
            \hdashline[1pt/3pt]
            BART($\CalD_{\sys}$) &$\textbf{53.70}$ &$\textbf{28.79}$ &$\textbf{50.81}$ \\
            \hline 
        \end{tabular}
\caption{Test set results on the SAMSum dataset. BART means fine-tuning BART on the original SAMSum. BART($\CalD_{\ke}$), BART($\CalD_{\rd}$) and BART($\CalD_{\ts}$) represent fine-tuning BART on the SAMSum with keywords, redundancy and topic annotation respectively. BART($\CalD_{\ke+\rd}$) represent fine-tuning BART on the SAMSum with keywords and redundancy annotations. $\CalD_{\sys}$ means the SAMSum with all three annotations.} \label{tab:ab_samsum}
\end{table}

\begin{table}[t]
\centering
        \begin{tabular}{l|lll}
            \hline
            \textbf{Model} & \textbf{R-1} & \textbf{R-2} & \textbf{R-L}   \\
            \hline
            \hline
            \multicolumn{4}{c}{\it Ours} \\
            \hline
            PGN  &48.34 &16.02 &23.49 \\
            \hdashline[1pt/3pt]
            PGN($\CalD_{\ke}$)  &50.22 &17.74 &24.11 \\
            PGN($\CalD_{\rd}$) &50.62 &16.86 &24.27 \\
            PGN($\CalD_{\ts}$) &48.59 &16.07 &24.05 \\
            \hdashline[1pt/3pt]
            PGN($\CalD_{\ke+\rd}$)  &50.74 &17.11 &24.52 \\
            PGN($\CalD_{\ke+\ts}$) &50.69 &16.83 &24.33 \\
            PGN($\CalD_{\rd+\ts}$) &50.70 &16.96 &24.38 \\
            \hdashline[1pt/3pt]
            PGN($\CalD_{\sys}$) &\textbf{50.91} &$\textbf{17.75}$  &$\textbf{24.59}$ \\
            \hline
        \end{tabular}
\caption{Test set results on the AMI dataset. PGN($\CalD_{\ke}$), PGN($\CalD_{\rd}$) and PGN($\CalD_{\ts}$) represent training PGN on the AMI with keywords, redundancy and topic annotation respectively.  PGN($\CalD_{\ke+\rd}$) represent training PGN on the AMI with both keywords and redundancy annotations.} \label{tab:ab_ami}
\end{table}

\section{Hyper-parameter Search Results}
Tables \ref{tab:samsum_ke} to \ref{tab:ami_ts} show the hyper-parameter search results.
Finally, for SAMSum \cite{samsum}, we set keywords extraction ratio $r_{\ke}$ to 15, similarity threshold $t_{\rd}$ to 0.99 and topic segmentation ratio $r_{\ts}$ to 15. for AMI \cite{ami}, $r_{\ke}$ is 4, $t_{\rd}$ is 0.95 and $r_{\ts}$ is 5. 

\begin{table}[!htb]
\centering
        \begin{tabular}{c|c|ccc}
            \hline
            \textbf{Model}&$\mathbf{r_{\ke}}$ & \textbf{R-1} & \textbf{R-2} & \textbf{R-L}   \\
            \hline
            BART($\CalD_{\ke}$) & 10  &52.17 &26.64 &48.34  \\
            BART($\CalD_{\ke}$)& \textbf{15}  &\textbf{53.43} &\textbf{28.03} &\textbf{49.93}   \\
            BART($\CalD_{\ke}$) & 20  &53.20 &28.01 &49.46  \\
            BART($\CalD_{\ke}$) & 25  &52.78 &27.35 &48.67   \\
            \hline 
        \end{tabular}
\caption{Test set results on the SAMSum dataset. BART($\CalD_{\ke}$) means fine-tuning BART on SAMSum with keywords annotation. $r_{\ke}$ means different keywords extraction ratios.} \label{tab:samsum_ke}
\end{table}

\begin{table}[!htb]
\centering
        \begin{tabular}{c|c|ccc}
            \hline
            \textbf{Model}  &$\mathbf{r_{\ke}}$ & \textbf{R-1} & \textbf{R-2} & \textbf{R-L}   \\
            \hline
            PGN($\CalD_{\ke}$)  &3  & 49.76	&16.03	&23.64\\
            PGN($\CalD_{\ke}$)  &\textbf{4}  &\textbf{50.22}	&\textbf{17.74}	&24.11   \\
            PGN($\CalD_{\ke}$)  &5  &49.63 	&16.71 	&23.88   \\
            PGN($\CalD_{\ke}$)  &6  &49.70 	&16.92 	&\textbf{24.42}    \\
            \hline 
        \end{tabular}
\caption{Test set results on the AMI dataset. PGN($\CalD_{\ke}$) means training PGN on AMI with keywords annotation. $r_{\ke}$ means different keywords extraction ratios.} \label{tab:ami_ke}
\end{table}

\begin{table}[!htb]
\centering
        \begin{tabular}{c|c|ccc}
            \hline
            \textbf{Model} & $\mathbf{t_{\rd}}$ & \textbf{R-1} & \textbf{R-2} & \textbf{R-L}   \\
            \hline
            BART($\CalD_{\rd}$) & 0.95 &52.29 &26.71 &48.53  \\
            BART($\CalD_{\rd}$) & 0.96 &53.20 &27.98 &49.68 \\
            BART($\CalD_{\rd}$) & 0.97 &52.17 &27.10 &48.34 \\
            BART($\CalD_{\rd}$) & 0.98 &53.29 &27.89 &\textbf{49.71} \\
            BART($\CalD_{\rd}$) & \textbf{0.99} &\textbf{53.39} &\textbf{28.01} &49.49 \\
            \hline 
        \end{tabular}
\caption{Test set results on the SAMSum dataset. BART($\CalD_{\rd}$) means fine-tuning BART on SAMSum with redundant utterances annotation. $t_{\rd}$ means different similarity thresholds.} \label{tab:samsum_td}
\end{table}

\begin{table}[t]
\centering
        \begin{tabular}{c|c|ccc}
            \hline
            \textbf{Model} &$\mathbf{t_{\rd}}$ & \textbf{R-1} & \textbf{R-2} & \textbf{R-L}   \\
            \hline
            PGN($\CalD_{\rd}$) &\textbf{0.95}	&\textbf{50.62}	&\textbf{16.86}	&24.27 \\
            PGN($\CalD_{\rd}$) &0.96	&49.68	&16.54	&\textbf{24.70} \\
            PGN($\CalD_{\rd}$) &0.97	&50.18	&16.12	&24.56 \\
            PGN($\CalD_{\rd}$) &0.98	&48.63	&15.17	&23.50 \\
            PGN($\CalD_{\rd}$) &0.99	&47.15	&13.94	&22.53 \\
            \hline 
        \end{tabular}
\caption{Test set results on the AMI dataset. PGN($\CalD_{\rd}$) means training PGN on AMI with redundant utterances annotation. $t_{\rd}$ means different similarity thresholds.} \label{tab:ami_td}
\end{table}

\begin{table}[t]
\centering
        \begin{tabular}{c|c|ccc}
            \hline
            \textbf{Model}  &$\mathbf{r_{\ts}}$ & \textbf{R-1} & \textbf{R-2} & \textbf{R-L}   \\
            \hline
            BART($\CalD_{\ts}$)  &10 &53.21 &27.38 &49.32  \\
            BART($\CalD_{\ts}$)  &\textbf{15}  &\textbf{53.34}  &\textbf{27.85}  &49.64 \\
            BART($\CalD_{\ts}$)  &20  &52.82  &27.34  &49.05 \\
            BART($\CalD_{\ts}$)  &25  &53.04  &27.49  &\textbf{49.70} \\
            \hline 
        \end{tabular}
\caption{Test set results on the SAMSum dataset. BART($\CalD_{\ts}$) means fine-tuning BART on SAMSum with topic annotation. $r_{\ts}$ means different topic segmentation ratios.} \label{tab:samsum_ts}
\end{table}

\begin{table}[t]
\centering
        \begin{tabular}{c|c|ccc}
            \hline
            \textbf{Model} &$\mathbf{r_{\ts}}$ & \textbf{R-1} & \textbf{R-2} & \textbf{R-L}   \\
            \hline
            PGN($\CalD_{\ts}$) &4 &49.39 &16.02 &23.89 \\
            PGN($\CalD_{\ts}$) &\textbf{5} &48.59	&\textbf{16.07}	&\textbf{24.05}\\
            PGN($\CalD_{\ts}$) &6 &\textbf{49.89}	&16.04	&23.01 \\
            PGN($\CalD_{\ts}$) &7 &49.37	&\textbf{16.07}	&23.46 \\
            \hline 
        \end{tabular}
\caption{Test set results on the AMI dataset. PGN($\CalD_{\ts}$) means training PGN on AMI with topic annotation. $r_{\ts}$ means different topic segmentation ratios.} \label{tab:ami_ts}
\end{table}

\end{document}